\documentclass[manuscript, screen]{acmart}

\AtBeginDocument{%
  }

\setcopyright{acmlicensed}
\copyrightyear{2024}
\acmYear{2024}
\acmDOI{XXXXXXX.XXXXXXX}

\usepackage{tabularray}
\usepackage{booktabs}
\usepackage{multirow}
\usepackage[normalem]{ulem}
\useunder{\uline}{\ul}{}

\begin{document}

\title{The State of Post-Hoc Local XAI Techniques for Image Processing: Challenges and Motivations}

\author{Rech Leong Tian Poh}
\affiliation{
    \institution{T{\"U}V S{\"U}D Asia Pacific}
    \country{Singapore}
}
\affiliation{
    \institution{School of Computing Science, University of Glasgow}
    \country{Scotland, United Kingdom}
}
\email{rech.leong@tuvsud.com}

\author{Sye Loong Keoh}
\affiliation{
    \institution{School of Computing Science, University of Glasgow}
    \country{Scotland, United Kingdom}
}
\email{syeloong.keoh@glasgow.ac.uk}

\author{Liying Li}
\affiliation{
    \institution{School of Computing Science, University of Glasgow}
    \city{Glasgow}
    \country{Scotland, United Kingdom}
}
\email{liying.li@glasgow.ac.uk}

\begin{abstract}
As complex AI systems further prove to be an integral part of our lives, a persistent and critical problem is the underlying black-box nature of such products and systems. In pursuit of productivity enhancements, one must not forget the need for various technology to boost the overall trustworthiness of such AI systems. One example, which is studied extensively in this work, is the domain of Explainable Artificial Intelligence (XAI). Research works in this scope are centred around the objective of making AI systems more transparent and interpretable, to further boost reliability and trust in using them. In this work, we discuss the various motivation for XAI and its approaches, the underlying challenges that XAI faces, and some open problems that we believe deserve further efforts to look into. We also provide a brief discussion of various XAI approaches for image processing, and finally discuss some future directions, to hopefully express and motivate the positive development of the XAI research space. 
\end{abstract}


\maketitle

\section{Introduction}
As AI-driven systems increasingly back and power high-stakes decision-making in various public domains such as healthcare, finance, criminal justice, and autonomous vehicles, the need for explainability becomes ever more prevalent and critical for end-users to take informed and accountable actions \cite{introcite1}. Explainable AI (XAI) is the research area that tackles the need for clear understanding on how a decision is made by an AI system, as typically there is a lack of trust that comes from the use of black-box AI systems and products. In many cases, human users are having difficulty deciphering the outcome of an AI system, this is especially important when the outcome conflicts with the human interpretation. With this, XAI aims to provide human-understandable information to provide insights into an AI system's behaviour and processes. 

There have been many industry applications such as autonomous vehicles (AV), medical, and aviation that have started to embed AI components into its operation, e.g., in the AV domain, accurate object detection is required to determine hazardous events and subsequently control the vehicles to avoid accidents from happening; while in the medical domain, many image diagonostic tools have been developed to detect cancer cells, tumours, etc. All these are applications safety-critical, hence demanding explainability to predict failure cases during the testing phase, and to diagnose unwanted situations where AI components are in use. With explainability techniques,  the manufacturers can better understand certain failure cases in which their interfacing AI components have failed, leading to potential damage to their respective stakeholders. This serves as an important step in refining their products while performing testing during its development lifecycle. The use of explainability on AI system can be beneficial for post-hoc investigation, e.g., to diagnose an accident that was caused by a fault from an AI component, thus deriving insights that contribute towards an understanding of causality between interfacing components and an overall accountability and traceability of the origins of the faults. In such a manner, underlying problems in AI components can be better understood and avoided or remediated in the future, so that any safety-related fatal accidents may be prevented. 

Explainability is as much of a human-centric problem, as it is a technical one. When we think about making AI systems explainable, it is important to first understand the target audience. Who are we making the explanations available to? Who are the ones that are going to request for such explanations? This begs the question: \lq\lq explainable to whom?\rq\rq \cite{human-XAI1}. The subject of the explanation typically governs the most effective way of describing the \textit{why} behind the decision-making process. Most research works target an algorithmic-centred approach \cite{shap,limepaper}, whereas much work needs to be done to translate this into meaningful information in the regulatory space \cite{aiact}. Explainability tends to be a multi-stakeholder problem which spans across the entire AI lifecycle, taking into consideration the demands of system engineers, domain experts, and end users. This brings about very different and distinct goals to the explanation task \cite{multistakes}. 

While AI practitioners and end users typically emphasise on the aspects of high performance, accuracy and reliability, a significant amount of regulatory pressure is impending as well. According to the European Parliament, people have a right to explainability whenever an AI system has a significant impact on their lives \cite{gdpr}. This means that any user of an AI system has the right to seek an explanation for how their data is being used or how the AI system operates, especially in situations where the AI system's decision has an impact on their lives. One such example is the hiring process, where AI components could be used to automate the recruitment process \cite{amazon}. The introduction of AI in such a process does not change the fact that applicants have the right to know the reasons for which their application was accepted or rejected. Additionally, the EU AI Act sees constant updates and signals an enforcement date in the very near future \cite{aiact}.

This paper aims to establish an understanding of the existing motivations for XAI, and tie them to some underlying challenges. We discuss the potential benefits of using XAI, not only to leverage upon to make informed technical decisions, but also the potential integration of various XAI elements to boost explainability. We conclude this survey with some insights, and future directions that the research areas can expand in. Our contributions in this work are as follows:
\begin{itemize}
    \item An overview of the overarching motivations for XAI, as well as their respective challenges faced.
    \item Further insights into the open problems faced by current XAI methodology.
    \item The potential future directions that XAI research space may take.
\end{itemize}

\section{Background} 

To provide some context, we categorise relevant papers and tabulate them in table \ref{table:survey_papers}. While each paper presents their own take on their respective key topics within XAI research, they can be broadly categorised based according to three type(s); \textit{Domain-Specific}, \textit{Human-Centric}, or \textit{Socio-Techinical}. The papers outlined here can present overlaps in terms of the categories they belong to, depending on their respective scope(s). We define the categories and describe them as the following:

\begin{itemize}
    \item \textbf{Domain-Specific} - Survey papers that focus specifically on the XAI approaches, demands, and studies done within a certain domain (i.e. Medical, Automotive etc.), aligning key objectives and advancements of XAI to the requirements of the respective key stakeholders (i.e. doctors, field experts).
    \item \textbf{Human-Centric} - Survey papers that have a concentrated effort on the human aspects of XAI, such as \textit{interpretability}, \textit{transparency}, and \textit{bias}. Such papers have the underlying objective(s) of highlighting the need for XAI to reinforce fundamental human rights such as \textit{privacy} and \textit{accountability}.
    \item \textbf{Socio-Technical} - Survey papers of this nature demonstrate the mapping between the technical elements of XAI such as \textit{accuracy}, \textit{consistency}, and \textit{robustness} to societal objectives of XAI such as \textit{transparency},  \textit{ease of use}, and \textit{user experience}.
\end{itemize}

\begin{table}[ht]
\centering
\begin{tabular}{@{}cccc@{}}
\cmidrule(l){2-4}
\multicolumn{1}{l}{} & \multicolumn{3}{c}{Category} \\ \midrule
\multicolumn{1}{c|}{Paper} & \multicolumn{1}{c|}{Domain-Specific} & \multicolumn{1}{c|}{Human-Centric} & Socio-Technical \\ \midrule
\multicolumn{1}{c|}{\cite{ROBLESCARRILLO2020101937, socialxai}} & \multicolumn{1}{c|}{} & \multicolumn{1}{c|}{$\checkmark$} & \\ \midrule
\multicolumn{1}{c|}{\multirow{2}{*}{\cite{doshivelez2017rigorous, doran, soc1, survey2018, holzinger2020, Molnar_2020, tech1, zhou2021evaluating, ras2021explainable, poeta2023conceptbasedexplainableartificialintelligence, zhao2023explainabilitylargelanguagemodels, electronics8080832,danilevsky-etal-2020-survey,burkart2020survey}}} & \multicolumn{1}{c|}{\multirow{2}{*}{}} & \multicolumn{1}{c|}{\multirow{2}{*}{}} & \multirow{2}{*}{$\checkmark$} \\
\multicolumn{1}{c|}{} & \multicolumn{1}{c|}{} & \multicolumn{1}{c|}{} & \\ \midrule
\multicolumn{1}{c|}{\cite{Pocevi_i_t__2020,8419428, medical, lucieri2020achievements}} & \multicolumn{1}{c|}{$\checkmark$} & \multicolumn{1}{c|}{$\checkmark$} & \\ \midrule
\multicolumn{1}{c|}{\cite{industry1,perspectives,nagahisarchoghaei_empirical_2023}} & \multicolumn{1}{c|}{$\checkmark$} & \multicolumn{1}{c|}{} & $\checkmark$ \\ \midrule
\multicolumn{1}{c|}{\multirow{2}{*}{\cite{le_benchmarking_2023,human-XAI1,introcite1,personexp, survey2023, mersha_explainable_2024, r_transparency_2024,ras2018explanation,tradeoff,soc4,fairsurvey,mueller2019explanation}}} & \multicolumn{1}{c|}{\multirow{2}{*}{}} & \multicolumn{1}{c|}{\multirow{2}{*}{$\checkmark$}} & \multirow{2}{*}{$\checkmark$} \\
\multicolumn{1}{c|}{} & \multicolumn{1}{c|}{} & \multicolumn{1}{c|}{} & \\ \bottomrule
\end{tabular}
\caption{Categorisation of Survey Papers According to their Respective Objective(s)}
\label{table:survey_papers}
\end{table}

To shed some light into the cited papers in Table \ref{table:survey_papers} where more than one category was assigned:

\begin{itemize}
    \item \textbf{Domain-Specific \& Human Centric} - Papers belonging to this particular mix of categories \cite{Pocevi_i_t__2020,8419428, medical, lucieri2020achievements} had a particular focus on the \textit{medical} sector. These papers prioritised the alignment of XAI technology at their respective point(s) of writing, to human rights requirements pertaining to patients seeking treatment, and clinicians using AI-enabled systems to aid in medical diagnosis. The general trend noticed here was the clearly high demand for XAI in the medical sector, where not only is safety and privacy extremely prioritised, but providing varying degrees of transparency to cater to various stakeholders is equally just as important. 
    \item \textbf{Domain-Specific \& Socio-Technical} - Papers belonging to this particular mix of categories \cite{industry1,perspectives,nagahisarchoghaei_empirical_2023} had a focus on the applicability of XAI onto various industrial contexts and use cases. They explored the readiness and maturity of XAI to meet stakeholder requirements. The general consensus and noticed trend was the need to focus on the needs and demands of various stakeholders during the entire development and operation process of an AI system or product, to the point of making it a key consideration during initial requirement gathering stage(s). The findings of papers in this category further reinforce the points made in \cite{multistakes}, placing emphasis on the target audience of generated explanations, and their respective user experience and expertise(s).
    \item \textbf{Human-Centric \& Socio-Technical} - Papers belonging to this mix of categories \cite{le_benchmarking_2023,human-XAI1,introcite1,personexp, survey2023, mersha_explainable_2024, r_transparency_2024,ras2018explanation,tradeoff,soc4,fairsurvey,mueller2019explanation} had a focus on a \textit{human-in-the-loop} approach towards XAI, placing particular emphasis on the need for \textit{Human-Centric} XAI technical development. Papers in this category aim to cater to the general audience, rather than specific groups of stakeholders as in the previous described category. 
\end{itemize}

Collectively, the papers mentioned here, along with the many others are a positive influence in the XAI research space, bringing many steps closer to achieving a unified framework for trustworthy and responsible AI.

The terminology, approaches, and perspectives of the papers collated in table \ref{table:survey_papers} are represented in the sections that follow, providing a comprehensive overview, insight, and discussion about the key research areas in XAI. We do not intend for this paper to present a broad coverage of all the research done in the XAI domain, but rather to serve as an initial starting point for readers to grasp the key concepts in XAI, with a focus on image processing. 

\section{Terminology in Explicable AI (XAI)}
This section outlines the terms commonly seen in the field of XAI, with regards to the output explanations generated by XAI algorithms.

\subsection{Explainability}

This refers to the inner workings and parameters of the model that lead to its decision-making ability(s). Such information is usually hidden in deep and complex neural networks, giving rise to their black-box nature.  Explainability can also represent the knowledge of what each node and/or attribute in a neural network, and their contribution or importance to the model's result, thus providing a form of justification for the decision(s) made. 

Explanations are therefore systematically generated outputs that provide more information which gives users an insight into an AI system's decision making process. Such information can take many forms, such as top-k features, features ranked in terms of influence, degree of correlation etc.

XAI approaches typically generate explanations in the form of graphs and/or heatmaps in order to aid the visualisation and human comprehension. In order to produce such information, XAI approaches leverage on probing techniques or surrogates \cite{shap,limepaper} to derive relational distributions between input and output, thus deducing some degree of causality.

\subsection{Interpretability}
While often used interchangeably with \emph{explainability}, interpretability refers more to the aspect of causality, measuring how accurate a machine learning model can associate or relate cause to effect, or in other words, the correlation from input to output. 

To describe an example to distinguish between explainability and interpretability, a flowchart of an AI system's decision making process may allow a practitioner to understand the inter-communication between the AI model and the rest of the system (interpretability), but does not provide sufficient information for an in-depth understanding of how and why a certain decision was made (explainability) \cite{explain}. 

\subsection{Trustworthiness}

Trustworthiness is commonly defined as the \lq\lq confidence of whether a model will act as intended when facing a given problem \cite{trust}.\rq\rq In other words, trust is built upon aligning user expectations to the outcomes produced by AI systems. Trust can also be a culmination of various aspects \cite{trust}:
\begin{enumerate}
    \item \textbf{Human Agency and Oversight}
    \begin{itemize}
        \item Fundamental Rights - Loss of personal data can often become a large risk that should be reduced and/or justified. Mechanisms should be put into place to receive external feedback regarding AI systems that potentially infringe on fundamental rights.
        \item Human Agency - Users should be able to make informed autonomous decisions regarding AI systems. They should be given the knowledge and tools to comprehend and interact with AI systems to a satisfactory degree and, where possible, be enabled to reasonably self-assess or challenge the system.
        \item Human Oversight - Human supervision and discretion should be established during use of an AI system, so as to ensure the ability to override a decision made by the AI system in the case of failures and unwanted results. The less oversight a human is allowed over an AI system, the more extensive testing and stricter governance is required.
    \end{itemize}
    \item \textbf{Technical Robustness and Safety}
    \begin{itemize}
        \item Resilience to Attack and Security - AI systems should be well protected against vulnerabilities that could allow them to be attacked by adversaries.
        \item Fallback Plan and General Safety - AI systems should have contingency plans in case of failures, both to safeguard data and functions that could dampen the safety of its users.
        \item Accuracy - AI systems should be able to make correct judgments and/or predictions. In the case of an inaccurate prediction, the system should be able to indicate how likely these errors are. 
        \item Reliability and Reproducibility - AI systems should be able to exhibit the same behaviours when given the same set of inputs repeatedly. They should also be able to work properly with a range of inputs and in a variety of situations, as long as they are within the scope of its intended design.
    \end{itemize}
    \item \textbf{Privacy and Data Governance}
    \begin{itemize}
        \item Privacy and Data Protection - This must be ensured throughout the system's entire lifecycle. In order to allow users to trust the data gathering process, the data collected must not be used to unlawfully or unfairly discriminate against them.
        \item Quality and Integrity of Data - Prior to training, gathered data needs to be rid of any socially constructed biases, inaccuracies, errors and mistakes. The training processes and datasets used must also be tested and documented at each step of the development lifecycle. 
        \item Access to Data - Data protocols that govern data access should be implemented. Such protocols should dictate data access rights and only allow qualified and competent personnel to access private data.
    \end{itemize}
    \item \textbf{Transparency}
    \begin{itemize}
        \item Traceability - The relevant decisions, data sets and processes that are made with regards to the AI system should be well labelled and documented to allow for traceability and increase in transparency.
        \item Explainability - When AI systems have significant impact on people's lives, it should be possible to demand a reasonable explanation of the decision making process.
        \item Communication - An AI system's capabilities and limitations should be communicated properly to its end-users appropriately according to its use-case. 
    \end{itemize}
    \item \textbf{Diversity, Non-Discrimination and Fairness}
    \begin{itemize}
        \item Avoidance of Unfair Bias - Any form of biases, whether in training or operation datasets should be alleviated by putting oversight processes in place to analyse and address the system's purpose, constraints, requirements and decisions in a clear and transparent manner. 
        \item Accessibility and Universal Design - AI systems should consider addressing the widest possible range of users and be inclusive to all societal groups.
        \item Stakeholder Participation - It is advised to consult stakeholders who may directly or indirectly be affected by the AI system throughout its lifecycle. 
    \end{itemize}
    \item \textbf{Societal and Environmental Well-Being}
    \begin{itemize}
        \item Sustainable and Environmentally Friendly AI - The lifecycle of the AI system should be one that is ecologically responsible and consider resource usage and energy consumption both during training and operation.
        \item Social Impact - The effects of AI systems on the social lives of its users must be carefully monitored and considered.
        \item Society and Democracy - Use of AI systems must be given careful and ample thought before being included into democratic processes such as political decision-making and electoral contexts.
    \end{itemize}
    \item \textbf{Accountability}
    \begin{itemize}
        \item Auditability - Evaluation both by internal and external auditors and the availability of their reports can contribute to the trustworthiness of the AI system. AI systems should be able to be independently audited when used in applications that affect fundamental rights, including safety-critical applications.
        \item Minimisation and Reporting of Negative Impacts - We should be able to report on actions or decisions that contribute to a certain system outcome and to react to the consequences of unintended outcomes. 
        \item Trade-Offs - Tensions may arise when working towards a trustworthy AI product while maintaining its intended functions. As such, any conflicts that arise should be identified and unless these trade-offs are ethically acceptable, the AI system should not proceed in that form. 
        \item Redress - When unjust adverse impacts occur, accessible mechanisms should be foreseen to ensure sufficient remedy.
    \end{itemize}
\end{enumerate}

Each of the aspects mentioned above can be built using various processes such as risk analysis, robustness tests, and appropriate chain of command within an organisation. Explainability and its tools are just one of the many possible approaches that an organisation can take to achieve a higher degree of trust in an AI system. Having sound explanations and interpretations of an AI system's decision making process will heavily contribute to its overall trustworthiness. 

\subsection{Repeatability}
In terms of model performance, the model's inference result should remain the same when given the exact same inputs under the exact same environment to ensure consistency and repeatablity. For example, if a job applicant reapplies to a firm, assuming the exact same position, applicant resume, job acceptance criteria and employment terms, the resultant outcome should remain the same. Likewise, in the context of healthcare, an AI-produced diagnosis should remain the same when repeatedly being provided with the same patient data as input. 

\subsection{Reproducibility}
Similarly, the model's inference result should be achievable when used by different practitioners under different environments. The same behaviour should be observed in its respective explanation, with a relationship that is representative of the AI model's decision-making process.

\subsection{Explanation Stability}
To achieve higher levels of explanation stability, an output explanation should stay relatively consistent under multiple executions or explanation generations. This is an observed shortcoming of perturbation-based explanations approaches such as \cite{limepaper}, where explanation outputs may differ under repeated explanation generations for the same inputs. 

\subsection{Causality \& Cotenability}

Causality refers to the degree to which the AI system's decision-making process can be described using relationships between inputs and outputs, either graphically or visually.

In order for an explanation to be logically sound, we must also consider its cotenability, that is, the co-dependence among the various features presented in the black-box model. For example, in order for an AI system to predict an applicant's eligibility for a loan, consideration for the applicant's income, marital status, and credit score must be done in conjunction and not in isolation from one another. In practice, feature co-dependence is usually observed in most black-box models.

Another simple example to describe this correlation would be the computation of one's Body Mass Index (BMI), where there is a co-dependence between weight and height, such that it is impossible to change the BMI result while holding both the height and weight constant, but it is possible to attain the same BMI result for a different set of height and weight measurements. This relationship is captured and described very well in \cite{causal}.

An ideal explanation would have considerations for both cotenability and causality, such that the explanation represents both of these aspects as a causal effect of intervention.

\subsection{Faithfulness}
Faithfulness applies to both model predictions and the generated explanations. Both model outputs and their explanations are expected to be faithful to the ground truth represented in training data. In other words, faithfulness can also be interpreted as the degree to which outputs are aligned to the facts being represented by features in input data. 

For example, an NLP (Natural Language Processing) model is deemed as faithful if its output prediction is exactly as intended based on its input and the corresponding assigned label. Likewise, the explanation produced is equally as faithful, if it accurately represents the reasoning or representation of words or tokens in the sentence, with relevance to the correct output label.

\section{Why Explainable AI?}
The recent hype of AI systems and tools such as ChatGPT, Dall-E, BlackBox.ai, etc has resulted in using AI systems to make decisions or generate supporting information and assets. This has helped boasting productivity boosts with high turnaround times. AI has also seen its use in various industry sectors, such as in autonomous vehicles, social networks, and medical systems. While being in awe of the sheer capability that AI products bring to the table, one must not forget that these products and systems are often black-box in nature. Decisions and information provided by such AI systems are often not explained, so the reasons and factors behind the generated results are typically unknown.

According to studies \cite{survey2018, survey2023, fairsurvey}, not all black-box AI systems need an explanation for why they take a certain decision because this might result in many drawbacks such as reduced efficiency and increased operation and development costs. Generally, studies agree that explainability is needed to a certain extent in most situations, other than in the following two situations \cite{two}:

\begin{itemize}
    \item The AI system generates results that are unacceptable, but do not lead to severe consequences.
    \item The AI system and its components has been studied in-depth and well-tested in practice. Additionally, supporting technical documentation should follow which clearly describes the underlying decision making process of the AI inference/system.
\end{itemize}

Such AI systems exhibit very low levels of risk. Some examples include postal code sorting and targeted advertisement systems. However in most cases, AI systems present higher levels of risk, especially when they are deployed in some safety-critical applications such as Autonomous Driving (AV) or medical diagnosis use-cases.

As covered by other studies \cite{perspectives}, the various works in XAI can be categorized into the following perspective groups:

\begin{itemize}
    \item Regulatory Pressure: As black-box AI systems see increased adoption in many aspects of our daily lives, the complexity of such systems can lead to some instances in which decisions and outputs produced by AI components can be inaccurate and/or unacceptable to stakeholders. In some cases, the repercussions include unwanted legal effects, leading to a new challenge and pressure from the regulatory perspective.

    The European Union's General Data Protection Regulation (GDPR)\cite{gdpr} is one such example of why XAI is needed from a regulatory perspective. These regulations dictate what is known as the\lq\lq right to explanation\rq\rq, by which a user has the rights to request for an explanation about any decision(s) made by an AI system, particularly if it considerably influences them \cite{right}. However, implementation of such regulations are never straightforward, often challenging, and lack the enabling technology with which explanations can be adequately provided\cite{requestright}.
   
    With the increased complexity and capability of Generative AI and Large Language Models (LLM), regulatory bodies have found it challenging to keep up to the pace of advancement. Even published regulations and standards constantly need to be updated, as more areas of applicability arises from such advanced AI products.
    \newline
    \item Industrial Adoption: The key challenges presented to the widespread adoption of complex and accurate AI products are mainly regulations and general user distrust \cite{industry1}. As such, less accurate but more interpretable models may be preferred in heavily-regulated industries, given the better adherence to current regulations \cite{industry1}. One key benefit of applying XAI techniques in such a situation, is to mitigate any further impacts and trade-offs between model interpretability and performance \cite{tradeoff}, thus creating an environment of increased adoption of complex AI products. However, a combination of XAI and complex AI products can increase development and deployment costs.
    \newline
    \item Technical Advancement: Limited or biased training data, outliers, adversarial data, and model overfitting are just some of the many reasons that contribute to the inappropriate and inaccurate results generated by black-box AI systems. As such, the decision making process and underlying learned features need to be better understood, especially when such AI-supported results have an impact on the safety and security of human lives. XAI techniques can be used here to understand, debug, and improve the black-box AI system to improve its overall robustness, enhance safety and privacy controls, establish user trust, and minimise any unwanted behaviour from bias or discrimination \cite{tech1}. XAI can also serve as a secondary quality gate when comparing models with similar performance, revealing other factors that may impact the overall decision \cite{tech2,tech3}. 
    \newline
    \item Societal Impact: In deep learning literature \cite{soc1,soc2,soc3}, it has been shown that making modifications to an image that cannot be detected by human eyes may lead an AI model into producing a wrong class label. Meanwhile, images that are completely unrecognizable to humans are instead very recognizable and produce high confidence levels in deep learning models \cite{soc3}. With that in mind, black-box AI systems may produce unfair decisions that were not apparent to the naked eye. For example, AI systems could be trained and developed using data that already exhibits human bias and prejudices, further amplifying them instead \cite{soc1}. As such, using XAI to produce explanations would help us understand the models' decisions and establish greater levels of trust. From the social perspective, there is a demand for the fairness of black-box AI systems' decisions, since this typically can not be measured or ensured by current approaches \cite{soc4}.
    
\end{itemize}

\begin{figure*}[!ht]
    \centering
    \includegraphics[scale=0.7]{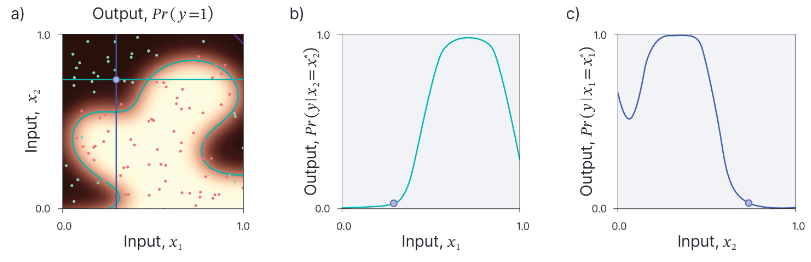}
    \caption{ICE: a) Understanding why the purple point is assigned to the negative class by considering what would happen if either feature $x_{1}$ (cyan line) or feature $x_{2}$ (purple line) was changed. b) The effect of changing feature $x_{1}$: This results in the point being classified as positive, if feature $x_{1}$ had a higher value. c) The effect of changing feature $x_{2}$: This results in the point being classified as positive if feature $x_{2}$ had a lower value. \protect\cite{ice}}
    \label{fig:ice}
    \Description{Illustration of ICE}
\end{figure*}

The list above is understandably non-exhaustive and presents some overlap between these perspectives. Prior studies express the agreement that these are the top level areas that XAI aims to tackle \cite{survey2023}. In this paper, we aim to provide a more comprehensive understanding of individual smaller problem sectors, and categorise all of these aspects in a larger scale.

\section{Post-Hoc Local Explainability Techniques}
Given the wide variety of XAI approaches that are designed for various scenarios \cite{survey2023}, a more common and realistic situation is the given access to an already trained complex model like a deep neural network, but we may not have access to its internal structure nor the data with which it was trained on. In this context, the model is considered black-box in nature, and we seek insight into how it makes decisions, using explainability techniques. This exact situation where explainability techniques are performed on existing models is referred to as post-hoc explanations.

Local post-hoc explanations evade the problem of trying to interpret an entire AI model by focusing on just explaining a particular subset of decisions. Many local post-hoc methods attempt to simply describe a specific narrow distribution of data points under consideration, rather than the entire decision-making process of a model.

\subsection{Individual Conditional Expectation (ICE)}

An \textit{individual conditional expectation} or Fig. \ref{fig:ice} shows an \textit{ICE} plot \cite{ice}, which takes an individual prediction and shows how it would change upon varying a single feature. Essentially, it answers the question of ``\emph{What if a feature had taken on another value?}" In terms of the input-output function, it examines the change in output for a single dimension of a given data point \ref{fig:ice}.

ICE plots have the disadvantage of only being able to examine a single feature at a time, thus being unable to take into account the relationships or correlations between two or more features. It is also possible that some combinations of input features may not happen in an operational environment.

\subsection{Counterfactual Explanations}
While ICE plots create insight into model behaviour by visualising the effect of changing one of the model inputs by a set amount, counterfactual explanations\cite{wachter2018counterfactual} manipulate multiple features but only consider the behaviour within the vicinity of a particular input for which to explain.

Counterfactual explanations are typically used within the context of classification. From the perspective of the end-user, it answers the question of ``\emph{What changes would I have to make for the model classification to be different?}" For example, in the case of a declined loan application, a counterfactual explanation might indicate that the loan decision would have been different if the application had two less credit cards and an extra \$5000 annual income. 

\begin{figure*}[!ht]
    \centering
    \includegraphics[scale=0.7]{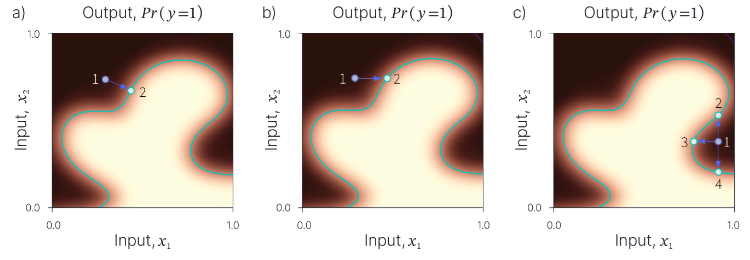}
    \caption{Counterfactual Explanations: a) Find out the explanation for which the data point (purple point 1) was classified negatively to determine what needs to be changed to yield the correct classification (cyan point 2). b) Seeking sparse counterfactual examples where  only a few features (i.e just feature $x_{1}$ are changed. c) There may be multiple potential ways to modify the input (brown point 1) to change the classification points. \protect\cite{ice} }
    \Description{Illustration of Counterfactuals}
    \label{fig:cf}
\end{figure*}

The main drawback of counterfactual explanations is that there may be multiple possible ways to change the model output by perturbing the features locally, so it is not clear which is the most useful. Additionally, since most approaches are based on the optimisation of complex and non-linear functions, finding existing and realistic counterfactuals within a predefined distance from the original data point can fail, but that does not mean that they do not exist.

Counterfactuals have two other drawbacks that were addressed by \cite{Dandl_2020}. First, it is preferable to have counterfactual examples in which only a small number of input features have changed. This is more understandable and practical in terms of taking remedial action(s). Second, it is important to ensure that the counterfactual example falls within a possible and realistic region of input space. With reference to the previous loan example, the decision could be made partly based on two different credit ratings, but they might be highly correlated. On the other hand, suggesting a change where one remains low, but the other is increased, is not particularly helpful since it is challenging to realise this in practice. In this regard, \cite{Dandl_2020} proposed adding a second term that penalised the counterfactual example if it is far from the training points. Another important modification was made in \cite{grath2018interpretable}, which allows the user to specify a weight for each input dimension that effectively penalises change more or less. This can be used to discourage the exploration of counterfactual examples where the changes made to inputs are not realistic, such as the proposal for a change to someone's age.

\subsection{Local Interpretable Model-Agnostic Explanations (LIME)}
\begin{figure}[!ht]
    \centering
    \includegraphics[scale=0.4]{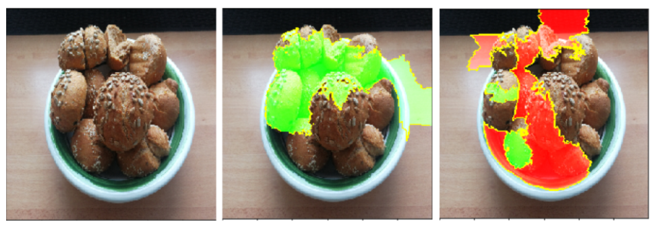}
    \caption{Left: Image of a bowl of bread. Middle and right: LIME explanations for the top 2 classes (bagel, strawberry) for image classification made by Google's Inception V3 neural network. \protect\cite{limepage}}
    \Description{Illustration of SHAP}
    \label{fig:lime}
\end{figure}

The authors of LIME \cite{limepaper} proposed an implementation of local surrogate models which are trained to approximate the predictions of a black-box model. Instead of training a global surrogate model, LIME focuses on training local surrogate models instead, to explain individual predictions. In order to produce the data required to train the surrogate model, LIME only uses the black box model probing it several times and then tests what happens to the resulting predictions when variations of the input are given to the black-box model. The goal is to understand why the model made a certain prediction. Over time, a new dataset is generated, consisting of perturbed samples and their corresponding predictions given by the black-box model. With this new dataset, LIME then trains an interpretable model (the local surrogate) which is weighted by the distance of the sampled instances to the instance of interest. The resulting learned model then becomes a good approximation of the predictions locally, but it does not have to be a good global approximation. Using the local surrogate model, interpreting it will then generate explanations for individual predictions. Fig \ref{fig:lime} illustrates an example of an output explanation generated by LIME.

\subsection{SHapley Additive exPlanations (SHAP)}
The goal of SHAP \cite{shap} is to explain the prediction of any given instance to a black-box model by computing the contribution of each feature to the resulting prediction. It uses Shapely Values \cite{shapvalue} to produce such explanations and has many variations of application for different types of Machine Learning models. 

\begin{figure}[!ht]
    \centering
    \includegraphics[scale=0.25]{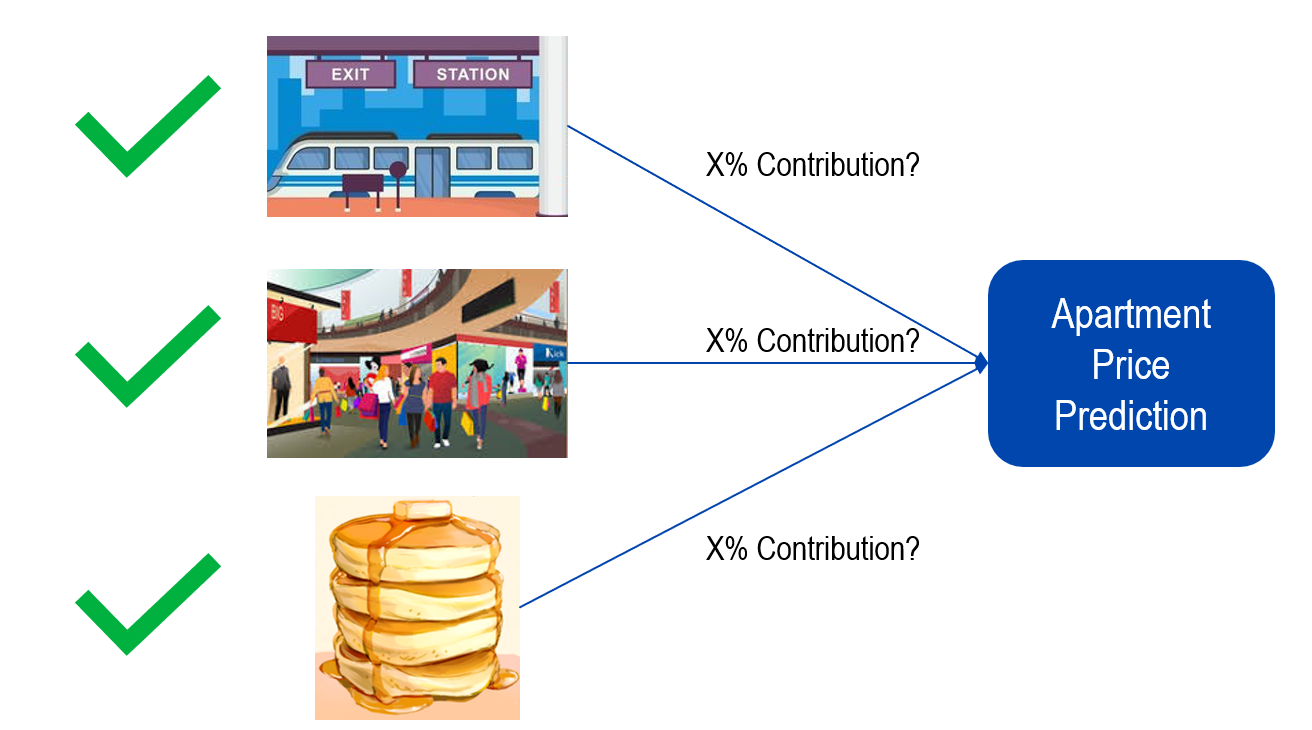}
    \caption{Illustration of Shapley Values}
    \Description{Illustration of Shapley Values}
    \label{fig:shapv}
\end{figure}

Shapley value was a method derived from coalitional game theory \cite{shapvalue}, which tells us how to fairly distribute the payout or loot among players of the game. This concept can be applied to machine learning, where each feature is considered to be a player and each player's payout is the influence or contribution that the individual features have on the final prediction as illustrated in Fig. \ref{fig:shapv}. Essentially, this produces Shapley values  as an interpretable additive feature attribution method, thus producing a simpler, surrogate explanation model that is more linear in nature. 

\begin{figure*}[!ht]
    \centering
    \includegraphics[scale=0.8]{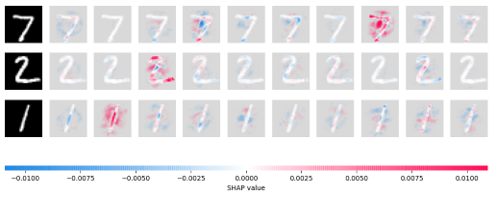}
    \caption{Illustration of SHAP Explanation \protect\cite{shappart}}
    \Description{Illustration of SHAP Explanation}
    \label{fig:shap}
\end{figure*}

Fig. \ref{fig:shap} shows an example in applying SHAP to explain a complex model. Colours are selected to represent positive and negative Shapley values. In this example, Magenta and Cyan represents the positive and negative Shapley values respectively. By observing the output explanation, one can derive that the higher concentrations of magenta spots on an MNIST digit represents a stronger correlation and argument that the model predicted it as such. For example, in Fig. \ref{fig:shap}, the first row (MNIST digit ``7") has a large concentration of magenta spots on the image along column 8 (otherwise representing output classification ``7").

\section{Challenges in XAI}
\begin{figure*}[!ht]
    \centering
    \includegraphics[scale=0.2]{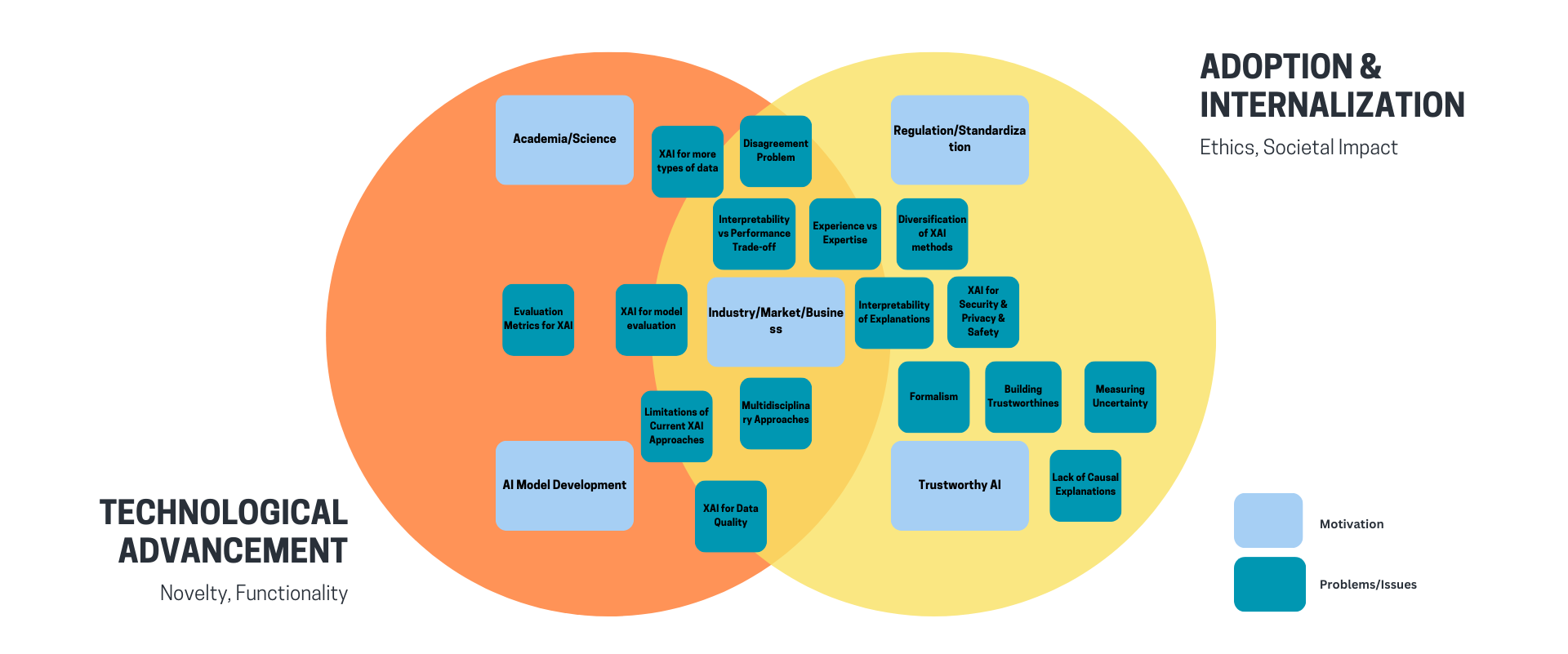}
    \caption{A Summary of the Motivations \& Challenges in XAI}
    \Description{Illustration of motivations \& challenges in XAI}
    \label{fig:xai_summary}
\end{figure*}

This section describes some of the challenges faced in the XAI domain in various sectors. A summary and illustration is shown in Fig. \ref{fig:xai_summary}, which illustrates and categorises the various high-level challenges in XAI, according to their respective motivations.

\subsection{Lack of Formalism}
A major problem in the XAI space is the lack of a systematic definition and quantification, or cohesive agreement on what explainability truly is. Various different definitions exist and even disagree with one another in terms of the level of depth or understanding required \cite{survey2023, multistakes}. Without a satisfactory definition of explainability and/or interpretability, it might not be possible to determine if new XAI approaches are better at explaining ML models and AI systems. 

To add on, not all XAI approaches can be used for all kinds of AI systems, due to their varying capabilities, limitations, and target audience(s) \cite{openxai, multistakes}. XAI approaches may also produce varying outputs and disagree with one another \cite{barr_disagreement_2023}, which is discussed later in section \ref{sec:disagreement_problem}.

Fundamentally speaking, the mentioned reasons are why there has yet to be a singular de-facto XAI approach or algorithm that can be applied to all AI systems.

As such, our view is that explainability needs to be more than simply explaining how an AI model came to a decision, but also need to consider the various stakeholders and target audience for which the explanation(s) are for. Elements of causality and what-if scenarios are great for the well-informed, but not necessarily to users who are not familiar with the technicalities and inner workings of AI systems.
\cite{8466590} suggests that there is a need for the consolidation of research works in order to build generic explainable frameworks for XAI that would better guide the development of end-to-end XAI approaches. As part of such frameworks, establishing of formalised rigorous evaluation metrics need to be considered as well. In light of the absence of an agreed definition of explainability and interpretibility, there is currently a lack of well established approach to evaluate XAI results. In such cases, the lack of ground truth explanations in often the largest challenge. Optimisation of such approaches also remains an open question, given that factors such as cost of predictability or run-time efficiency are still not well understood. 

There have been found to be two main evaluation metrics groups \cite{Molnar_2020}: 
\begin{itemize}
    \item Objective evaluation metrics - Reliant on quantifiable, mathematical metrics
    \item Human-centric evaluation metrics - Reliant on user studies and experimentation
\end{itemize}

Given the pretext that XAI largely remains a multi-stakeholder demand \cite{multistakes}, further work is needed towards evaluating the performance and establishing objective metrics for evaluating XAI approaches in various contexts and for different user groups \cite{arrieta_explainable_2020}. In the case of human-centric evaluations, there is a need for the development and adoption of effective designs for user studies and subjective explanation evaluation measures to help establish some agreed criteria on human-centric evaluations, this making comparisons between explanations easier \cite{zhou2021evaluating}. In the case of objective evaluations, previous works mainly focused on attribution-based explanations, such as feature importance \cite{shap,limepaper}. In order to work towards a more robust means of evaluating explanations, a larger variation of explanation elements and the integration of requirements from the various user-groups need to be considered as well. Additionally, there is a demand for more objective evaluations that measure other properties of explainability, such as interpretability and fairness of the explanations. 

In an ideal situation, both evaluation groups need to be integrated for a comprehensive evaluation, in order to understand the magnitude of the contribution of each evaluation metric on the overall analysis. In this manner, each explanation and respective evaluation is done based on the task and user group, which ultimately will be the best approach towards building a model-agnostic explanation evaluation framework \cite{electronics8080832}.

\subsection{Interpretability of Explanations}

Depending on the nature and the application of an AI system, the types of users who get exposed to them can vary (e.g. domain experts, data scientists, key decision makers, non-experts etc.) User experience and expertise would therefore be a confounding factor that directly affects the level of depth and complexity expected from an output generated by XAI algorithms \cite{doshivelez2017rigorous}. One of the key challenges towards enhanced interpretability of explanations is the lack of consideration towards bridging the gap between user expertise and depth of explanations\cite{ras2018explanation,ras2021explainable}.

For example, a study done in \cite{mueller2019explanation} highlights that previous works in XAI for expert systems generally did not take into account the knowledge and abilities of users. Additionally, the key objectives and contexts of the various stakeholders and users were not defined. In an effort to address this, works such as \cite{personexp} have discussed that identifying the users' objectives and keeping up with the agile requirements involves the collection of data from such users. It is also fundamental to develop approaches to manage and detect changes to these user objectives, and the need to adapt towards end-users. 

In \cite{burkart2020survey}, it was mentioned that Deep Learning models often use complex concepts that are unintelligible and often challenging to predict outcomes. Therefore, using such black-box systems would require contextually-aware explanations that can accurately explain a decision, while also being interpretable to the users (e.g. clinician or medical expert). \cite{lucieri2020achievements} suggests that we can do the following:

\begin{itemize}
    \item Examine the role human-understandable information represented in Deep Learning models.
    \item Analyse the features used by the Deep Learning models in predicting correct decisions, based on incorrect reasoning.
    \item Have an understanding of the model's concepts, to reduce reliability concerns and develop trust when deploying the AI system, via active stakeholder engagement and user manuals to boost familiarity with the AI system(s).
    \item Consider the various stakeholders and their expectations of the AI system.
\end{itemize}

In summary, it is crucial to tailor explanations towards user experience and expertise. Explanations should be catered differently to different users under different contexts \cite{royal}. It is also essential to clearly define the objectives of users, systems, and explanations alike. To achieve this, frequent stakeholder engagement is fundamental.

\subsection{Implementation: Complexity vs Accuracy Trade-Offs}

There seem to be an underlying notion that complex models provided more accurate outputs, but that is not necessarily correct \cite{rudin2019stop}. \cite{arrieta2019explainable} explores model interpretability in a situation where performance is coupled with model complexity. It suggests that XAI techniques could help in minimising the trade-off between model complexity and its accuracy. However, \cite{8419428} questions the factors that determines such a trade-off. The authors of \cite{8419428} have highlighted the importance of discussing such trade-offs with end-users, so that they are made aware of the potential risks of mis-classification or model opacity. 

Another point that needs further consideration is the dilemma of approximation, that models need to be explained in sufficient detail and in a manner that matches the target audience, while keeping in mind that explanations need to reflect the model without oversimplifying its essential features \cite{arrieta2019explainable}. Even though studying such trade-offs is needed, it is tough to proceed without the standardisation of metrics to evaluate the quality of explanations \cite{danilevsky-etal-2020-survey}.

A possible solution for the trade-off was suggested in \cite{holzinger2020}, which is to develop fully transparent models throughout the entire process of the AI system development life cycle, while providing local and global explanations. As such, this led to the use of methods that embed learning capabilities to develop accurate models and their respective representations. The suggested methods should also be able to describe these representations using effective natural language that is consistent with human understanding and logical reasoning \cite{holzinger2020}.

\subsection{Diversification of XAI Approaches}

While there are some overlaps between XAI algorithms, each one seems to be addressing a different question \cite{lucieri2020achievements}. According to \cite{8466590}, a combination of various methods to obtain more detailed explanations is rarely considered. Rather than using each XAI algorithm separately, we should investigate how we can use them as basic components that can be interlinked and synergised to develop more innovative approaches \cite{8466590}. Furthermore, this could help to provide explanations and related information in simple human-interpretable language \cite{ras2018explanation}. In that regard, initial efforts were cited in \cite{lucieri2020achievements}, where authors proposed a model that could provide both visual relevance and textual explanations \cite{park2018multimodal}. Multi-modal explanations have also been explored in \cite{chen2022rex}, with the proposal of a new framework which encompasses XAI algorithms based on a wide range of images and vocabulary. Generated textual explanations are paired with their corresponding visual regions in the image. In doing so, much better logical reasoning and interpretability were achieved compared to some state-of-the-art explanation models. These studies suggest opportunities for future research, with the aim of enhancing both intepretability and accuracy \cite{lucieri2020achievements}.

\subsection{Lack of Research in Causal Explanations}

Causal justifications Creating causal explanations for AI systems that is, explaining why the algorithms generated the predictions rather than how they did so can aid in improving human comprehension \cite{Pocevi_i_t__2020}. Furthermore, causal explanations make models more resilient to adversarial attacks and increase in significance when incorporated into decision-making processes \cite{Molnar_2020}. But there may be contradictions between causality and performance prediction \cite{Molnar_2020}. 

Using a loan application as an example, causality would deduce that a higher individual's income had a positive impact on loan approval. On the other hand, this might differ from performance prediction, which considers the accuracy and correctness of the model under various situations. Causality may not consider for all the types of situations that affect model performance, hence producing contradictions or inconsistencies in some cases.

Causal explanation, given its enhanced representation of logical reasoning, is anticipated to be at the forefront of ML research and will become an integral part of XAI literature \cite{Molnar_2020}. According to a recent survey on causal interpretation for ML \cite{causal}, it seems that the absence of ground truth for causal explanations and verification of causal relationships makes the evaluation of causal interpretability even more challenging. Therefore, more research is needed towards both the development and the evaluation of causal interpretability models \cite{causal}.

\subsection{Limitations of Current XAI Approaches}

The majority of XAI works is mainly on image and textual data. While XAI research do consider other types of data, they were noted to receive less attention \cite{rojat2021explainable}. 

The use of visualisation techniques to transform non-image data into images creates opportunities to unravel explanations through saliency maps \cite{shap,limepaper}. However, this should not be the only way forward for non-image and non-textual data. For example, existing XAI approaches for image and text need to be adjusted for use with graph data \cite{9875989}. Additionally, XAI approaches for image data, such as saliency maps, might need expert knowledge to be understood when applied to time series data \cite{rojat2021explainable}.

AI systems with several interconnecting AI-enabled components tend to use a wide variety of data types, so explainability approaches that can handle such heterogeneity of information will prove to be more in demand \cite{doi:10.1148/ryai.2020190043}. To give an example, such systems can simulate diagnostic processes of clinicians in the medical domain, where both images and physical readings are utilised in decision making \cite{lucieri2020achievements}. Consequently, the diagnostic effectiveness of the systems and the overall explainability of the system and its processes can be enhanced \cite{lucieri2020achievements}.

The overall scalability of XAI algorithms with regard to resources needed and overall practicality, was also noted to be a challenge \cite{8419428}. For example, for each situation that requires an explanation, a separate creation of a local model using LIME needs to be invoked \cite{limepaper}. Similarly, when computing Shapely values \cite{shap, shapvalue}, all possible combinations of feature inputs must be considered when computing variable contributions. Hence, computations can be rather costly for explanation domains that present a lot of variables.

While XAI algorithms such as SHAP \cite{shap} and LIME \cite{limepaper} are becoming increasingly more prevalent with their model-agnostic and visually interpretable explanation capabilities, more work needs to go into the compatibility with other model types and algorithms, e.g.,  both of these methods do not natively support object detection models, and require some tweaking in order to derive output explanations on a per-class basis \cite{moradi2023modelagnostic}.

In that regard, feature dependence also presents problems in attribution and extrapolation \cite{Molnar_2020}. If features are tightly correlated, presenting the effects of feature importance becomes challenging. For explanation analysis which relies on feature permutations such as SHAP and LIME, these feature correlations will be broken and result in data points outside of the regular distribution, therefore creating misleading explanations. Similarly, heatmap explanations present a lack of clarity regarding feature correlations. For example, such heatmaps only highlight the significance of specific pixels of an image, but do not indicate the correlation between various features in the image, such as objects or scenes \cite{Samek_2019}.

Authors in \cite{murdoch2019} has also explored some challenges for post-hoc explainability models. According to the authors, there is a difficulty in determining the format or combination of formats of explanations that will accurately describe a model's decision-making process. Additionally, there is uncertainty if current explanation methods are sufficient to capture model behaviour, or if novel methods are still needed. 

Lastly, some research directions have been suggested in \cite{dazeley2021explainable} to deal with the challenges faced by perturbation-based methods. For example, there is an underlying problem of finding the optimal set of perturbations of the inputs, since sampling all perturbations is non-exhaustive. The development of cross-domain applications of perturbation techniques will also benefit from empirical studies that compare perturbations across various data types \cite{dazeley2021explainable}.

\subsection{XAI for Model Evaluation \& Debugging}
Another challenge is if post-hoc explanations methods identify learned relationships by the model that practitioners know to be incorrect, is it possible that practitioners fix these relationships learned and increase the predictive accuracy? Further research in post-hoc explanations can help exploit prior knowledge to improve the predictive accuracy of the models.

\section{Open Problems in XAI}
In this section, we outline the various open problems in XAI that are very current and heavily under exploration, but constantly face setbacks and thus hamper its overall development. We discuss the literature and other related works, with a focus on the same overall objective of achieving various aspects of explanation quality.

\subsection{Disagreement Problem in XAI Algorithms}
\label{sec:disagreement_problem}
 Situations where explanations produced by multiple XAI approaches disagree with one another \cite{krishna2022disagreement} were highlighted. For example, they may disagree in the top-K most significant features. In such situations, practitioners might find it challenging to make informed decisions about which explanation and which XAI approach to rely on. It is uncertain how common the disagreement problem is in practice, given little to no study on the frequency of such disagreements, and for which types of XAI algorithms they may occur in. This problem is further enhanced by the inherent problem with the stability of output explanations, especially in the case of post-hoc attribution-based XAI methods \cite{shap,limepaper,survey2023}. 

Practitioners must address such disagreements head-on if and when they arise, because failing to do so could force them to rely on misleading explanations, leading to disastrous results. Examples of such consequences include the adoption and utility of AI models and systems that are racially biased, believing in inaccurate model predictions, and suggesting less-than-ideal recommendations to users.

Works such as \cite{krishna2022disagreement, comparexai, openxai} have taken initial approaches to quantify and justify disagreements among explanation outputs, accounting for data types that include image, textual, and tabular data. To quantify such disagreements, \textit{Alignment Metrics} are used. A list of alignment metrics can be found in the list below.

\begin{itemize}
    \item \textit{Feature Agreement (FA)} metric computes the fraction of top-K features that are common between a given post-hoc explanation and the corresponding ground truth explanation.
     \item \textit{Rank Agreement (RA)} metric measures the fraction of top-K features that are not only common between a given post hoc explanation and the corresponding ground truth explanation, but also have the same position in the respective rank orders. 
     \item \textit{Sign Agreement (SA)} metric computes the fraction of top-K features that are not only common between a given post hoc explanation and the corresponding ground truth explanation, but also share the same polarity (direction of contribution) in both the explanations. 
     \item \textit{Signed Rank Agreement (SRA)} metric computes the fraction of top-K features that are not only common between a given post hoc explanation and the corresponding ground truth explanation, but also share the same feature attribution polarity (direction of contribution) and position (rank) in both the explanations. 
     \item \textit{Rank Correlation (RC)} metric computes the Spearman’s rank correlation coefficient to measure the agreement between feature rankings provided by a given post hoc explanation and the corresponding ground truth explanation. 
     \item \textit{Pairwise Rank Agreement (PRA)} metric captures if the relative ordering of every pair of features is the same for a given post hoc explanation as well as the corresponding ground truth explanation i.e., if feature A is more important than B according to one explanation, then the same should be true for the other explanation. More specifically, this metric computes the fraction of feature pairs for which the relative ordering is the same between the two explanations.
\end{itemize}

\subsection{Evaluation Metrics for XAI Algorithms}
To better understand and assess the various XAI algorithms, some studies have adopted evaluation metrics and even developed novel approaches to benchmark and categorise the performance of XAI algorithms \cite{openxai, Consisxai, krishna2022disagreement, comparexai}. In addition to \textit{Alignment Metrics}, the following were found to be adopted for the purpose of evaluating XAI algorithms \cite{openxai}:

\begin{itemize}
    \item \textit{Prediction Gap on Important Feature Perturbation (PGI)} which measures the difference in prediction probability that results from perturbing the features deemed as influential by a given post-hoc explanation.
     \item \textit{Prediction Gap on Unimportant Feature Perturbation (PGU)} which measures the difference in prediction probability that results from perturbing the features deemed as unimportant by a given post-hoc explanation. 
     \item \textit{Relative Input Stability (RIS)}, \textit{Relative Representation Stability (RRS)}, \textit{Relative Output Stability (ROS)} which measures the maximum change in explanation, relative to the changes in the inputs, internal representations learned by the model, and output prediction probabilities respectively. 
\end{itemize}

In order to compare XAI algorithms with one another and formulate a benchmark for a specific use-case, the authors of \cite{openxai, comparexai} have come up with a list of tasks for each XAI algorithm to be assessed to accomplish. Each task is suited towards a quality characteristic such as \textit{Fidelity}, \textit{Fragility}, \textit{Stability}, \textit{Fairness}, and \textit{Consistency}. These tasks were expressed as \textit{Tests}, which presented the notion of functional testing practices adopted towards the evaluation of XAI algorithms. A list of some of the tasks are as follows:

\begin{itemize}
    \item \textbf{Fidelity - Does the algorithms's output reflect the underlying model? :} Test whether features of different importance are represented correctly, and test the effect of feature product on local explanations.
    \item \textbf{Fragility - Is the explanation result susceptible to malicious corruption? :} Attempt adversarial attacks to lower the importance of specific features.
    \item \textbf{Stability: Is the algorithm's output too sensitive towards slight changes in the data or the model? :} Test effect of data distribution and noisy input data.
    \item \textbf{Stress - Can the algorithm explain models trained on larger datasets or big data? :} Testing if the XAI algorithm is sensitive to a high number of word tokens (NLP Task), and test detection of dummy pixels in MNIST dataset.
\end{itemize}

\begin{figure*}[!ht]
    \centering
    \includegraphics [width=140mm]{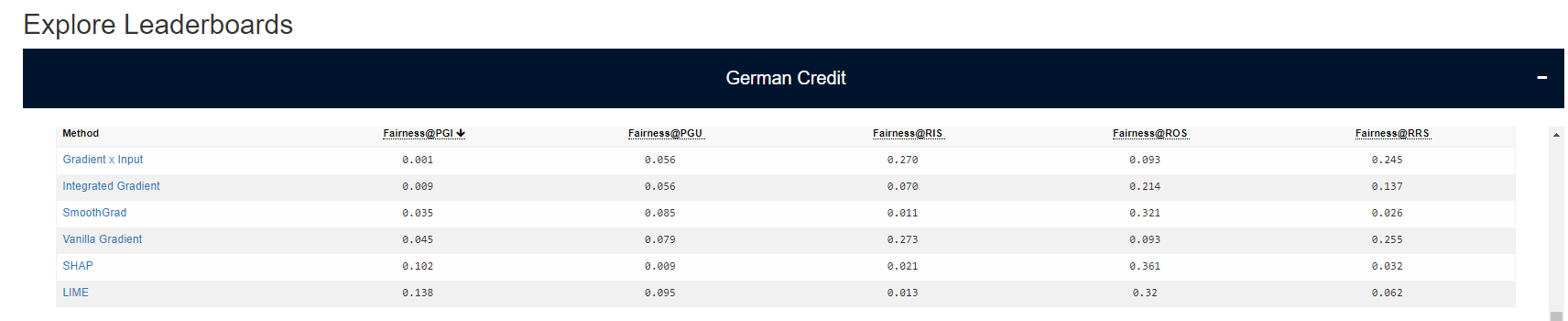}
    \caption{OpenXAI Leaderboard Snippet \protect\cite{openxai}}
    \Description{OpenXAI Leaderboard Snippet}
    \label{fig:openxai}
\end{figure*}

Once the tests have been completed, scores are determined for each XAI algorithm and categorised into their respective quality metric(s). The scores are then tabulated and interpreted as a leaderboard, thus creating a benchmark for subsequent XAI algorithms to be tested. An example can be found in Fig. \ref{fig:openxai}, extracted from \cite{openxai}.

\begin{figure*}[!ht]
    \centering
    \includegraphics[scale=0.6]{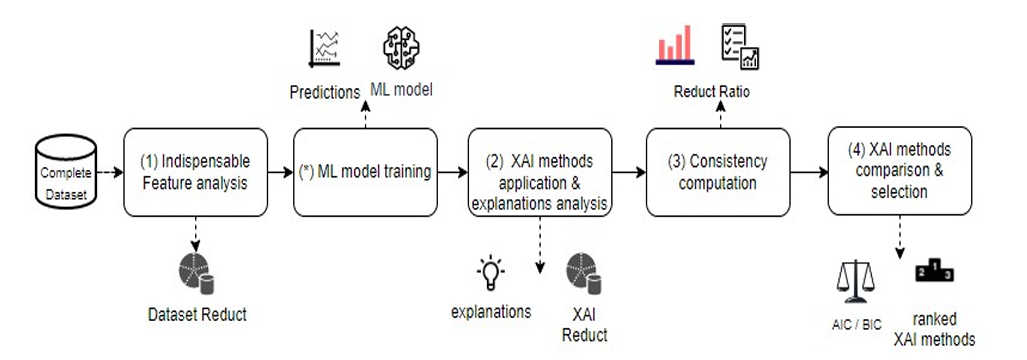}
    \caption{ConsisXAI Overview \protect\cite{Consisxai}}
    \Description{ConsisXAI Overview}
    \label{fig:consisxai_overview}
\end{figure*}

While the evaluation metrics and benchmarking approaches served their purpose, it is not always practical to make assumptions during the evaluation of ground truth explanations. For example, the authors of \cite{Consisxai} have identified the lack of a foundational ground truth, since most evaluation methodologies do not currently adopt a human-in-the-loop approach. As such, the authors of \cite{Consisxai} have used \textit{Feature Selection} approaches to reduce datasets into a set of indispensable features, as a better representation of ground truth. Thereafter, the XAI algorithms are evaluated against each other, but only for the purpose of quantifying the \textit{Consistency} of each XAI algorithm output against the dataset reducts. An illustration of this approach is shown in Fig. \ref{fig:consisxai_overview}, extracted from \cite{Consisxai}. 

The main pipeline and stages are broken down as follows:
\begin{itemize}
    \item Analysis of a given dataset to identify a set of features that are considered indispensable for the prediction process (ground truth extraction).
    \item Extracting the features that are actually used by the ML model, based on the explanations proposed by different XAI methods.
    \item Computing the consistency value for the feature set suggested by each XAI method.
    \item Comparing different XAI methods based on their consistency ratios according to the model selection metrics proposed.
\end{itemize}

In cases where the resultant consistency ratios were too close to make a decision, model selection metrics \cite{aicbic} were used to better identify and select the best XAI algorithms for each particular use-case. 

\subsection{Disagreement Problem in XAI Evaluation Metrics}

While there are ongoing works to develop metrics to evaluate XAI algorithms and approaches \cite{comparexai,openxai,Consisxai}, as well as works to address such disagreements in explanation outputs \cite{krishna2022disagreement}, the problem does not end there. Such evaluation metrics have also been found to have disagreements among themselves, with authors mentioning that this should serve as a wake-up call to the entire XAI community \cite{barr_disagreement_2023}. 

Authors of \cite{barr_disagreement_2023} performed experiments using XAI methods on a range of public empirial datasets, using the tabular data to create classification models. The goal was to find a set of explanations that was deemed to be the most faithful, using the set of XAI evaluation metrics selected. Across the experiments, the ranked correlations showed little consensus on the notion of faithfulness in the explanations, and the authors claimed that this would leave end users without the required tools to make informed decisions of their XAI method selection and benchmark.

We believe that more work needs to be done to explore this subtopic, to further assess XAI evaluation metrics for other kinds of purposes other than faithfulness, such as robustness and consistency, to further understand the areas in which XAI evaluation metrics may disagree with one another. 

\section{Our Thoughts and Future Directions}
With all the relevant works to further enhance the capabilities and the reliability of XAI algorithms, we propose some potential directions that the XAI community can look into, to better meet the needs of the various stakeholders and target users of XAI algorithms.

\subsection{Multi-Modal or Intra-Model XAI Approaches}
Firstly, we see potential in the use of \textit{Mutli-Modal} and \textit{Intra-Model} XAI approaches, to further enhance the human interpretability and the predictive capabilities of XAI algorithms. This can be achieved in the following ways:

\begin{itemize}
    \item \textit{Intra-Modal} - A combination of existing XAI algorithm components or outputs of the same type, for the enhancement of XAI algorithms' explanation capabilities and human interpretability. This could refer to the combinations of similar output explanations, or similar approaches to generate explanations.
    \item \textit{Multi-Modal} - A combination of two or more types of XAI output types (e.g. Image + Text), to better express the insights that can be derived from various types of XAI outputs, to enhance human interpretability. An example of this is the use of both \textit{Visual} output from SHAP or LIME, and \textit{Natural Language} output from a \textit{Vision-Language Models} such as Contrastive Language–Image Pre-training (CLIP) \cite{radford2021learningtransferablevisualmodels}. Such modelling techniques are studied in \cite{bordes2024introductionvisionlanguagemodeling}.
\end{itemize}

We theorise that the utility of the above-mentioned approaches will further enhance human interpretability of XAI algorithms and explanations, in a way that reduces the reliability on human expertise, while also boosting the predictive capabilities of XAI algorithms. Such approaches have also been suggested in works such as \cite{survey2023} and \cite{mersha_explainable_2024}, but at the current time of writing, we have yet to see any works that progressed in this direction.

\subsection{Enhancement of XAI Evaluation Metrics}
While there have been multiple works that have expanded on evaluation metrics for XAI algorithms \cite{Consisxai,comparexai,openxai,le_benchmarking_2023}, few of these works actually focused on a human-in-the-loop approach. We advocate that such an approach is a key factor that contributes to the overall quality of XAI algorithms and their outputs, since ultimately, such outputs must be interpretable by various stakeholders, while also allowing researchers to better draw correlations between metric selection and factors that contribute towards better stakeholder satisfaction. This is particularly important for domains of higher levels of expertise, where even human judgment can vary. In such cases, a higher degree of explainability may be required from XAI algorithms, thus warranting stricter evaluation, verification and validation measures and benchmarks.

Furthermore, metrics that have been studied in works such as \cite{openxai,Consisxai,barr_disagreement_2023} have proven their purpose for their various objectives, but we believe that there is potential for combinations of such metrics to achieve an even greater level of depth of evaluations.

\subsection{Towards Better Understanding of Ground Truth and Stakeholder Requirements}
To date, regulations such as \cite{aiact} have faced challenges pertaining to requirements for explainability and the relevant techniques. While there is a mandate for the \textit{``right to explainability"}\cite{requestright}, there is little to no specificity as to how such a degree of explainability can be achieved. The solution to such a problem is not as straightforward as we might think, therefore encouraging practitioners and developers alike to better understand the requirements of their stakeholders \cite{multistakes}, as well as what exactly is deemed as an accurate representation of the ground truth. For example, in very generic use-cases, something as simple as a flow chart could be sufficient to represent the decision-making process of an AI model. However in more complex and safety-critical use-cases, that is clearly insufficient, and further insights need to be derived for predictive maintenance, preventive measures, and remedial actions \cite{Pocevi_i_t__2020,8419428,medical}. One can interpret such an approach in a similar manner as the \textit{Risk Analysis} \cite{schwerdtner2020risk, tbiasex} process for typical software development, where a certain appropriate measure is taken for each and every risk identified, scaling accordingly with their respective severity and impact(s). Such an approach is studied in \cite{rex}, although used for a different objective. In some specific cases where domain expertise is involved, such as medical, aviation, and construction domains, stakeholder requirements are aplenty and the ground truth is often difficult to ascertain \cite{medical}. As such, stricter explainability approaches should be implemented, in accordance with the higher stakes \cite{multistakes}.  

\section{Conclusion}
To summarize this study, we covered the baseline for the XAI domain in terms of the various motivations and challenges. We provided a brief overview of post-hoc local XAI methods, then discussed the open problems and some relevant research works in the XAI domain, highlighting the advancements that have led the XAI community one step closer towards attaining overall explanation reliability. To conclude the study, we have also included our personal insights on certain topics, proposing further directions in which XAI studies can look into for the better advancement of the XAI domain.

\bibliographystyle{ACM-Reference-Format}
\bibliography{references}

\end{document}